\documentclass[letterpaper, 10 pt, conference]{ieeeconf}  

\IEEEoverridecommandlockouts                              
                                                          
\usepackage{epsfig} 
\usepackage{mathptmx} 
\usepackage{times} 
\usepackage{amssymb}  

\usepackage[linesnumbered,ruled,vlined]{algorithm2e}
\usepackage{amsmath}
\usepackage{multirow}

\usepackage{graphicx} 
\usepackage{caption}
\usepackage{subcaption}

\usepackage{textcomp}
\usepackage{mathtools}
\usepackage{changes}
\usepackage{xcolor}
\usepackage{color, soul}
\usepackage{booktabs}
\usepackage{xstring}
\usepackage{hhline}
\usepackage{siunitx}
\usepackage{comment}
\usepackage{physics}
\usepackage{tikz}
\usepackage{cite}
\usepackage{hyperref}

\usepackage{cleveref}

\usepackage{multirow}

\title{\LARGE \bf
eCDT: Event Clustering for \\Simultaneous Feature Detection and Tracking
}

\author{Sumin Hu$^{1}$, Yeeun Kim$^{1}$, Hyungtae Lim$^{1}$, Alex Junho Lee$^{2}$,  and Hyun Myung$^{1*}$, \textit{Senior Member, IEEE}
\thanks{$^{1}$Sumin Hu, Yeeun Kim, Hyungtae Lim, and Hyun Myung are with the School of Electrical Engineering, Korea Advanced Institute of Science and Technology (KAIST), Daejeon, 34141, South Korea. {\tt\small \{2minus1, yeeunk, shapelim\}@kaist.ac.kr}}%
\thanks{$^{2}$Alex Junho Lee is with the Department of Civil and Environmental Engineering, KAIST, Daejeon, 34141, South Korea. {\tt\small alex\_jhlee@kaist.ac.kr}}%
\thanks{*The corresponding author: Prof. Hyun Myung}%
\thanks{This work was supported by BK21 FOUR and STRADVISION}%
}

\begin{document}

\captionsetup[figure]{labelformat={default},labelsep=period,name={fig.}}
\captionsetup[table]{labelformat={default},labelsep=period,name={table.}}

\maketitle
\thispagestyle{empty}
\pagestyle{empty}

\begin{abstract}

Contrary to other standard cameras, event cameras interpret the world in an entirely different manner; as a collection of asynchronous events. Despite event camera's unique data output, many event feature detection and tracking algorithms have shown significant progress by making detours to frame-based data representations. This paper questions the need to do so and proposes a novel event data-friendly method that achieve simultaneous feature detection and tracking, called \textit{event Clustering-based Detection and Tracking~(eCDT)}.
Our method employs a novel clustering method, named as \textit{k-NN Classifier-based Spatial Clustering and Applications with Noise (KCSCAN)}, to cluster adjacent polarity events to retrieve event trajectories. 
With the aid of a \textit{Head and Tail Descriptor Matching} process, event clusters that reappear in a different polarity are continually tracked, elongating the feature tracks. Thanks to our clustering approach in spatio-temporal space, our method automatically solves feature detection and feature tracking simultaneously.
Also, eCDT can extract feature tracks at any frequency with an adjustable time window, which does not corrupt the high temporal resolution of the original event data. Our method achieves 30\% better feature tracking ages compared with the state-of-the-art approach while also having a low error approximately equal to it. 



\end{abstract}

\section{INTRODUCTION}

\begin{figure*}
    \captionsetup{font=footnotesize}
    \centering
    \includegraphics[width=\textwidth]{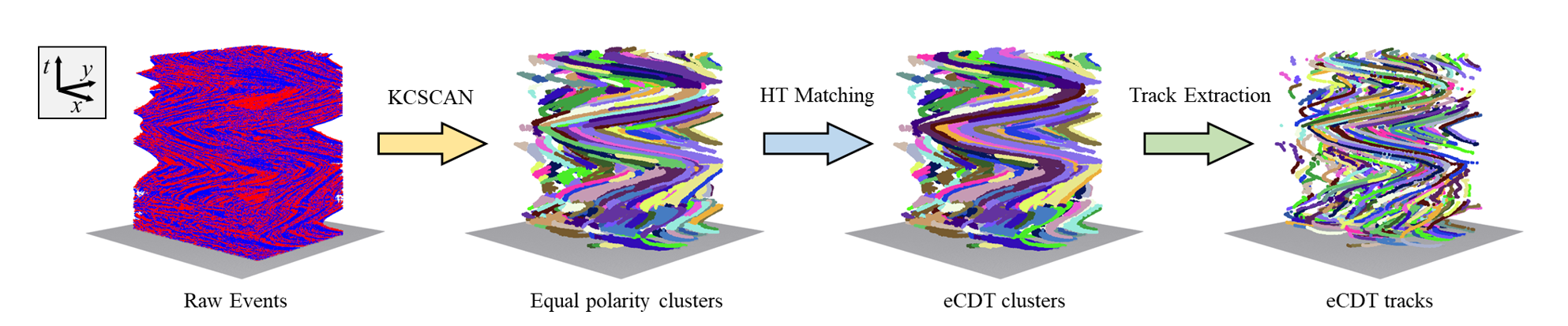}
    \caption{A full overview of our method called \textit{event Clustering-based Detection and Tracking~(eCDT)}. Raw events in 3D with their polarity information are processed by KCSCAN. Events with the same polarity are clustered. \textit{Head and Tail Descriptor Matching (HT Matching)} elongates the clusters that cannot be continually tracked due to polarity inversion. Finally, the eCDT tracks are extracted from the eCDT clusters.}
    \label{fig:pipeline}
\end{figure*}

\setlength{\abovecaptionskip}{1pt} 
\setlength{\belowcaptionskip}{1pt}

An event camera is a neuromorphic sensor that mimics the functions of an eye \cite{surveypaper}; it detects the intensity changes in its pixels that are output as signals called events. These events are either \textit{on} or \textit{off} events caused by a positive or negative change of intensity, respectively. Also, since these events are created asynchronously and independently per pixel, event cameras are granted the following three merits \cite{surveypaper}: a) robustness against highly dynamic situations, b) high temporal resolution, and c) low power requirements. Noticing the potentials of event cameras, the computer vision community has been paying increased attention towards this sensor.

Contrary to most camera variants, such as thermal cameras~\cite{ICRA2019_vivid}, which are frame-based, event cameras are pixel-based. To be more specific, whereas frame-based cameras output frames at synchronous time intervals, event cameras output each pixel data asynchronously. Due to this difference, conventional computer vision algorithms are not compatible with event cameras. As a result, a common method of coping with event data is to create event frames via accumulating pixel data within a time window. Another widely utilized data representation is the surface of active events (SAE), where frames are created with pixel intensity values representing how recent an event occurred in that pixel. Overall, while the frame-based representations of event data enables the adoption of vast precedent computer vision techniques, the data representation is suboptimal \cite{surveypaper, Evaluation_2021_journal}. In other words, these algorithms are not fully exploiting the advantages of event cameras.

In our paper, we diverge from frame-based representations and explore the capabilities of clustering as a feature detection and tracking algorithm. 

Our key contributions are as follows:
\begin{itemize}
  \item We formulate a novel clustering method called k-NN Classifier-based Spatial Clustering Application with Noise (KCSCAN) that clusters adjacent equi-polarity events but also segregates the clusters by opposing polarity events, thus resulting in a spatio-temporal clustering of features.
  
  \item By finding continual cluster tracks using cluster descriptors, event Clustering-based Feature Detection and Tracking (eCDT) shows about 30 percent longer tracks compared to the state-of-the-art algorithm (HASTE).

  \item Additionally, since KCSCAN clusters in a spatio-temporal space, it simultaneously performs feature detection and tracking, automatically detecting features and tracking them at the same time.
  
\end{itemize}

We believe this approach could be a stepping stone to more variants of algorithms that is more fitted to event-based data.

\section{RELATED WORKS}

\subsection{Clustering as a Means of Object Detection and Tracking}
Object detection and tracking for event cameras were explored early on. In 2006, \cite{cctv} proposed a vehicle tracking system on highways by clustering events that were triggered by passing cars. They took advantage of the fact that a stationary event camera setup only triggers events that correspond to moving objects. The stationary event camera setup has been extensively used in further studies by \cite{bolten2019application,cvpr-w2010}. In contrast to the aforementioned works, \cite{iros2018_multiTargetClustering, ICRA2020_intrusion} developed methods that even worked when the event camera was not stationary. Nonetheless, clustering was a key component of detecting and tracking the objects. 

\subsection{Predefined Shape Detection and Tracking}
\cite{tro2012Ni} identifies the state of grippers by tracking the positions of pre-defined left and right gripper locations in the image by using an Iterative Closest Point (ICP) algorithm. \cite{shape1_circle} tracks the location of a ball for iCub, an event-driven humanoid robot. With the shape of the object being circular, the authors extends the Hough-based circle detection algorithm for event cameras by using optical flow. In \cite{shape1_line}, line features are detected by getting intuitions from the Line Segment Detection (LSD) method. It calculates the orientation of the spatial derivative for each pixel and clusters events of similar orientation to form lines. 

\subsection{Corner Detection}
Feature detection and tracking of specific points of interest, such as corners, has shown progress in recent years with the adoption of a different type of data representation, which is called the Surface of Active Events (SAE). It imitates a frame-like representation where its pixel `intensity' values are higher with more recent events triggered at that pixel. Leveraging SAE's frame-like nature, many corner detection algorithms have been developed \cite{alzugaray2018asynchronous,mueggler2017fast,eHarris,FA-Harris} by taking intuitions from common corner detectors \cite{HARRIS,FAST}. 

\cite{eHarris} was the first to introduce a method of corner detection for event cameras. Their method, called eHarris, is an adaptation of the original HARRIS algorithm \cite{HARRIS}. Following up, \cite{mueggler2017fast} was the first to adapt the original Features from Accelerated Segment Test (FAST) algorithm \cite{FAST} and called their method eFAST. Both eHarris and eFAST detected the corner points by first converting the events into frame-based images by the SAE representation. Based on the two approaches suggested above, an abundant variation of approaches were suggested such as \cite {alzugaray2018asynchronous, FA-Harris}. As a variation of eFAST \cite{mueggler2017fast}, \cite{alzugaray2018asynchronous} can detect corners with obtuse angles. Also, this variant utilized a modified version of the SAE. \cite{FA-Harris} is a further development of \cite{alzugaray2018asynchronous}, but also accompanying the eHarris algorithm to process corner events. In contrary to these classical methods, \cite{SILC} proposes a method that trains a random forest decision tree that assess if an event is a corner point or not. This method utilizes another vairant of SAE called \textit{Speed-invariant Surface of Active Events}. As summarized above, corner point detection is mostly done in the SAE representation.

\subsection{Feature Point Tracking}
A variety of feature tracking algorithms were also published \cite{HASTE4, ACE2, wacv2020, gehrig2020eklt}. First of all, the eKLT \cite{gehrig2020eklt} tracks features using both frames and events. Harris corners are detected from the frames and a patch around the corner is used to create a template, which is tracked with events. \cite{ACE2} approaches feature tracking by calculating the corner point descriptor and finding corner points in adjacent regions of the most similar descriptor. \cite{wacv2020} is a method that tracks features using template batch matching. The templates are generated by the motion compensated event frame along a Bézier curve. This creates a clearer image to track. At each timestamp, the events are stacked along the curve and creates a \textit{Motion Compensated Event Frame}. Similarly, \textit{multi-Hypothesis Asynchronous Speeded-up Tracking of Events} (HASTE) \cite{HASTE4} tracks its points by using template matching. In this work, multiple hypotheses are given according to the direction of movement: translation and rotation. With the combination of movements, the likelihood of each hypothesis is calculated and whenever a hypothesis overtakes the previously prevalent hypothesis, the unit movement corresponding to the new hypothesis is considered to be detected. With each new hypothesis becoming the null hypothesis, the template is updated and the tracking algorithm continues this process.

\section{eCDT: event Clustering-based \\Detection and Tracking}

\begin{figure}
    \captionsetup{font=footnotesize}
    \centering
    \begin{subfigure}[b]{0.24\textwidth}
        \includegraphics[width=\linewidth]{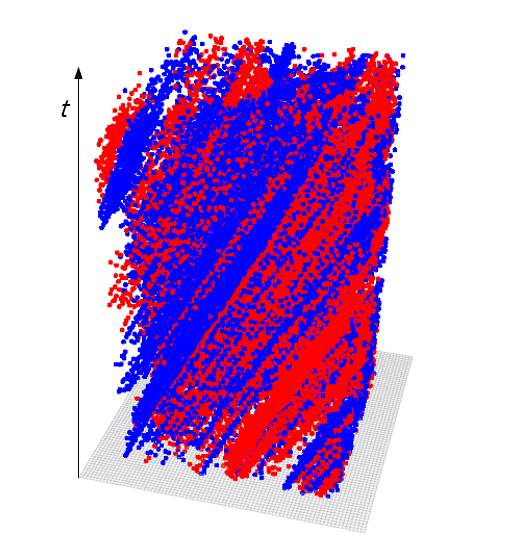}
        \caption{Oblique event view}
    \end{subfigure}%
    \begin{subfigure}[b]{0.24\textwidth}
        \includegraphics[width=\linewidth]{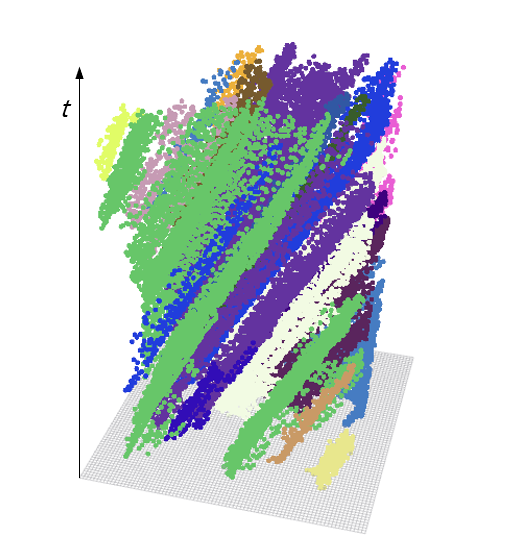}
        \caption{Oblique cluster view}
    \end{subfigure}%
    \vfill
    \begin{subfigure}[b]{0.24\textwidth}
        \includegraphics[width=\linewidth]{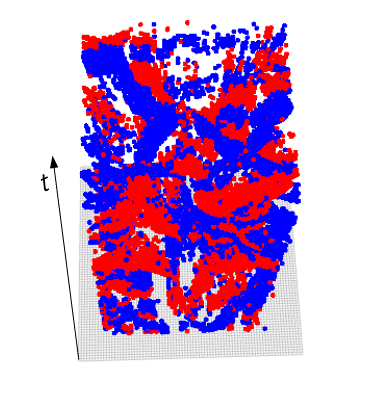}
        \caption{Clear top view}
    \end{subfigure}%
    \begin{subfigure}[b]{0.24\textwidth}
        \includegraphics[width=\linewidth]{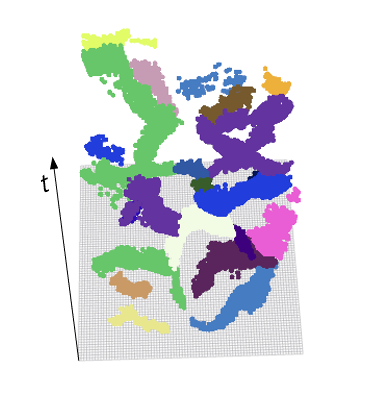}
        \caption{Cluster top view}
    \end{subfigure}
    \caption{Intuition behind \textit{event Clustering-based Detection and Tracking~(eCDT)}. (a) displays events in 3D with their polarity information. (b) displays the clustered events. (c) and (d) are top views of (a) and (b), respectively, that reveal the edge structures. 
    }
    \label{fig1:MainImg}
\end{figure}



\begin{figure}[h]%
    \captionsetup{font=footnotesize}
    \centering
    \subfloat[\centering Raw polarity points]{{\includegraphics[width=3.5cm]{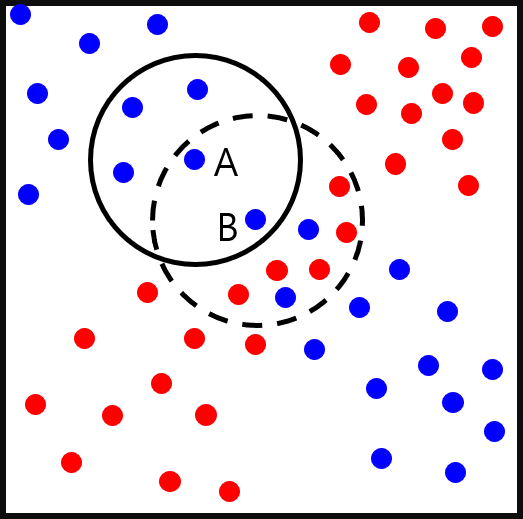} }}%
    \qquad
    \subfloat[\centering Clustered points]{{\includegraphics[width=3.5cm]{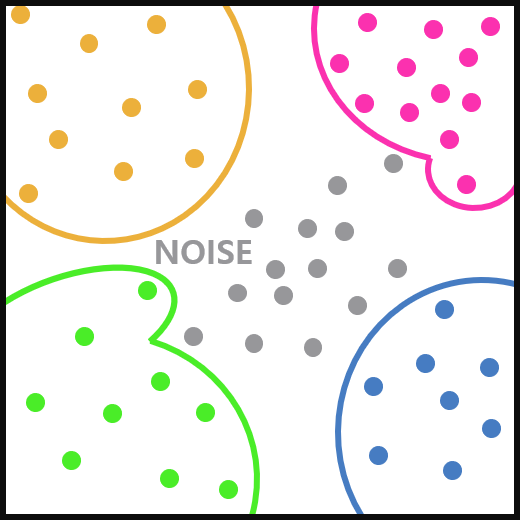} }}%
    \caption[clustering method explained with 2D illustration]{A 2D demonstration of boundary-assisted clustering. (a) Points in the image of red and blue polarity can be separated by opposite polarity points acting as barriers. Point \textit{A} is a core point with points of equal polarity nearby. However, point \textit{B} is classified as noise since neighboring points are not of equal polarity. (b) Resulting clusters are shown.}%
    \label{fig:2dClusteringDemo}
\end{figure}

\subsection{Intuitive Explanation of the Approach} \label{sec:intuition}
As mentioned previously, an event is triggered at a certain pixel ($x,y$) and time ($t$) if there is a (logarithmic) brightness change, due to moving edges, i.e., intensity gradients, in the image plane, that reaches a predefined threshold value $C$. With each pixel independently outputting events, the event data is a collection of these events: $ E_N = \{e_1, e_2, ..., e_N\}$, where ${e}_i = (x_i, y_i, t_i, p_i)$ is the $i$-th event representation that consists of the $x,y$ pixel coordinates, the time $t$, the polarity $p$, and $N$ is the total number of events.

In its raw form, i.e. the spatio-temporal space, as shown in Fig. \ref{fig1:MainImg}(a), event data are unintelligible, as stated by \cite{Gallego17ral_angular}. Therefore, motion compensation studies \cite{gallego2018unifying,Gallego17ral_angular} have shown that warping the events according to the trajectory of the edges leads to a crisp and clear revelation of the edge structures as shown in Fig. \ref{fig1:MainImg}(c). 
Fig. \ref{fig1:MainImg}(c) shows that by warping events according to their trajectory, events are clearly more visible by being aggregated with their own polarities. Therefore, motion compensation tries to maximize the image variance between the two polarities. 

In addition to their contribution, in our work, we focus on another key observation: edge patterns of different polarities create a boundary separating clusters of events. In Fig. \ref{fig1:MainImg}(b) and (d), we see clusters are visually separated by cluster of opposite polarities despite being close distanced. Therefore, neighboring opposite polarity events act as a cluster boundary that divides the spatio-temporal space. Referring to Fig. \ref{fig1:MainImg}(a), we notice the edge structures are actually manifested as a space dividing plane, which is what separates one cluster from another. Hence if we could cluster neighboring events of equal polarity bounded by opposite polarity clusters, we could modestly cluster events from the same edge and subsequently retrieve the trajectory. 



\subsection{KCSCAN: a Novel Means of Detection and Tracking}\label{sec:algorithm}



We now describe the crux of our proposed method. Our method depends on the observation that spatio-temporal adjacency of equal polarity events can function as a measure to discern whether events were triggered by the same edge structure or not. Thanks to the high temporal resolution of event cameras, a nearest neighbor approach to track objects or detected event features (e.g. corners) has been already prevalently practiced in this field \cite{bolten2019application, cctv, Evaluation_2021_journal, SILC, cvpr-w2010, iros2018_multiTargetClustering, ICRA2020_intrusion}.

\begin{figure*}[t!]
    \captionsetup{font=footnotesize}
    \centering
    \begin{subfigure}[b]{0.25\textwidth}
        \includegraphics[width=\linewidth]{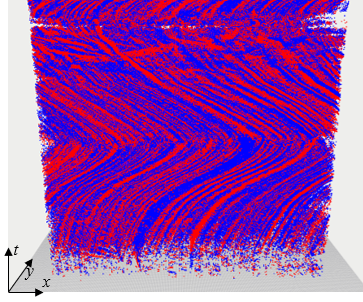}
        \caption{}
    \end{subfigure}%
    \hspace{5mm}
    \begin{subfigure}[b]{0.25\textwidth}
        \includegraphics[width=\linewidth]{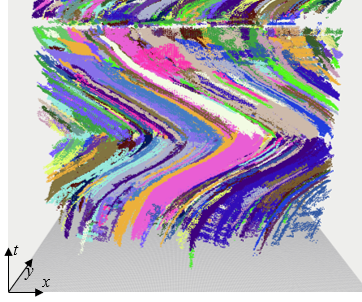}
        \caption{}
    \end{subfigure}%
    \hspace{5mm}
    \begin{subfigure}[b]{0.25\textwidth}
        \includegraphics[width=\linewidth]{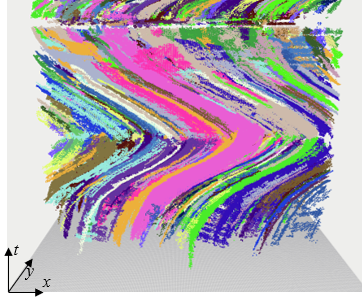}
        \caption{}
    \end{subfigure}%
    
    
    \caption[]{ (a) Raw events of 5 seconds are shown. Red indicates \textit{on} events and blue \textit{off} events. We can clearly notice a sudden change of direction in the middle from right to left. We also notice that the change of direction results in the change in polarity despite being from the same edge feature. (b) Clusters resulting from eCDT without the head and tail descriptor module show that the polarity inversion due to a sudden direction change results in discontinuous feature clustering. The pink cluster in the middle does not continue up, but starts as a new cluster in white. (c) eCDT clusters shows that longer tracking is possible. The head and tail descriptor match has lengthened the tracking for the pink cluster in the middle. A set of colors are randomly assigned to each cluster to differentiate between nearby clusters (best viewed in color).}
    \label{fig:3DVisualization}
\end{figure*}

Therefore, with the assumption that close equal polarity points are triggered from the same edge, we adopt Density-based Spatial Clustering of Apllications with Noise (DBSCAN) \cite{paper_DBSCAN} as a baseline method of clustering equal polarities. 
For completeness, we breifly describe DBSCAN. It is a non-parametric clustering algorithm that classifies points into a cluster if they are closely packed together. Clusters consists of points that are classified as core points when there are more than a predefined number of points $n$ within a predefined radius $r$. Neighboring points are also evaluated in an expansive manner, which therefore groups points of closely situated core points into a single cluster. 

However, DBSCAN alone cannot discern equal polarity clusters that are close but separated by opposite polarities. Therefore, we implement a subtle idea that recognizes opposite polarity events as borders when performing event clustering. We demonstrate the idea with the aid of some illustrations in 2D space. As shown in Fig. \ref{fig:2dClusteringDemo}(a), there are red points distributed from the bottom left corner to the upper right corner. Similarly, blue points are distributed from the bottom right corner to the upper left corner. Points of each color are positioned in a way that separates the differently colored points. By assessing if a point is neighboring any opposite polarity points, we can detect boundaries to a cluster. Therefore, with barriers separating the target polarity points, the 2D points should be classified as shown in Fig. \ref{fig:2dClusteringDemo}(b). Also, points that do not belong to any cluster are considered noise.

To implement this idea, we use a k-NN classifier that discerns a point's \textit{purity} by votes. More formally, assume we identified the \textit{k}-nearest neighbors $E_k$ of a query point $e_q = (x_q,y_q,t_q,p_q)$, where $E_k \subseteq E_N$. Then, 
\begin{equation}
    E_k = \{(x_1,y_1,t_1, p_1), (x_2,y_2,t_2, p_2), ...  ,(x_k,y_k,t_k, p_k)\}.
\end{equation}
We can define the purity score $\Phi_q$ for a query point $e_q$ as
\begin{equation}
    \Phi_q = \frac{1}{k}\sum_{i = 1}^{k}{\delta_{p_q,p_i}},
\end{equation}
where $\delta$ denotes the Kronecker Delta function. The purity score basically measures how pure, i.e., homogeneous, its surrounding is. With a minimum purity score predefined as $\Phi_{min}$, $e_q$ is a core point if $\Phi_q \geq \Phi_{min}$. So, referring back to Fig. \ref{fig:2dClusteringDemo}(a), assuming a purity score $\Phi_{min} = 1$, we notice that point $A$'s 4-nearest neighbors (excluding itself) are of equal polarity, making $A$ a core point. On the other hand, the 4-nearest neighbors of point $B$ have different polarity points, making $B$ just noise as shown in Fig. \ref{fig:2dClusteringDemo}(b). To prevent situations where k-NN are absurdly far away due to the sparsity of event data, we bound our k-NN search to a radius $r$. The algorithm that determines a core event is given in \texttt{Algorithm} \ref{alg1}. The iterative approach to expand the clusters with core points is adopted from DBSCAN, where neighboring core points have the same cluster index. This novel clustering method that uses a k-NN classifier for its core point evaluation is named as k-NN Classifier-based Spatial Clustering Applications with Noise (KCSCAN).

\begin{algorithm}[b]
\caption{eCDT Core Point Evaluation}
\label{alg1}
\SetAlgoLined
\SetKwInput{KwInput}{Input}
\SetKwInOut{KwOutput}{Output}
\DontPrintSemicolon
    \KwInput{ $\#$ of neighbors $k$, \ radius $r$,\ min purity $\Phi_{min}$ }
    \KwOutput{ is\_CorePoint, neighbors $N_k$ }
\SetKwFunction{FSub}{CorePoint}{}
\SetKwProg{Fn}{def}{:}{}
\Fn{ \FSub{$e_q$} }
{
    $N_k \gets \{\} $ \\
    $\Phi \gets$ 1 \\
    $E_k \gets kNN\_within\_radius(k,r)$ \\
    
    \For{$p_n \in E_k$}
    {
        \If{$p_n \neq p_q$}
        {
            $\Phi \gets \Phi-\frac{1}{k}$ \\
        }
        $N_k \gets  N_k  + \{p_n\}$ \\
        \If{$\Phi < \Phi_{min}$}
        {
            \KwRet $false$, \{\} \\
        }
    }

}
\KwRet $true, N_k$ \\
\end{algorithm}

\subsection{Head and Tail Descriptor Matching}\label{sec:descriptor}

As shown in Figs. \ref{fig:3DVisualization}(b) and \ref{fig:3DVisualization}(c), clustering equal polarities show weakness when there is a sudden change of direction that inverts the polarity of events created by the same edge. Therefore, we create a cluster descriptor that enables continuous tracking despite inverted polarities. 

Since we already recognize equal polarity cluster features with our clustering mechanism, data association between two clusters is imperative. Again, exploiting the spatio-temporal neighboring characteristic of events, events from the same edge will have spawned from a spatial position it disappeared, except for cases when features are lost because they left the camera's field of view. Thus, we only need to search the proximity of space where a cluster ended to recapture the lost features. The temporal search limit is the search time (s). Therefore, we formulate the problem as finding matching descriptors in a spatio-temporal space near where a cluster terminates.

In situations of sudden movement, the polarity inverts, whereas the shape created by the edge remains constant in a short time window. To elaborate, the small time interval where motion direction changes becomes a situation where no events are created and the position and visual shape of a feature practically remains intact. Therefore, the problem only requires matching the shape and position of a terminating cluster and newly created adjacent ones. Since the cluster's temporal boundary conditions, i.e. `head' or `tail' of a temporal trail of events, are only required to match clusters, we create cluster head descriptors to find preceding clusters from the same edge and cluster tail descriptors to find future clusters, originating from the same edge.  

\begin{figure}[t!]
    \captionsetup{font=footnotesize}
    \centering
    \begin{subfigure}[b]{0.1\textwidth}
        \centering
        \includegraphics[width=\textwidth]{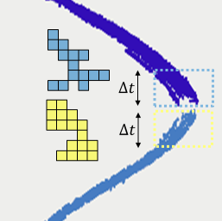}
        \caption[]{}%
    \end{subfigure}
    \hfill
    \begin{subfigure}[b]{0.1\textwidth}  
        \centering 
        \includegraphics[width=\textwidth]{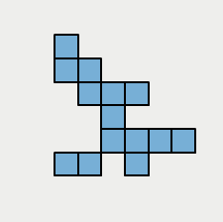}
        \caption[]{}%
    \end{subfigure}
    \hfill
    \begin{subfigure}[b]{0.1\textwidth}   
        \centering 
        \includegraphics[width=\textwidth]{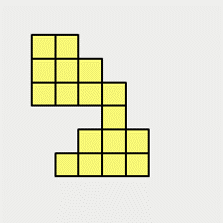}
        \caption[]{}%
    \end{subfigure}
    \hfill
    \begin{subfigure}[b]{0.1\textwidth}   
        \centering 
        \includegraphics[width=\textwidth]{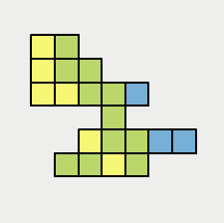}
        \caption[]{}%
    \end{subfigure}
    \caption[]%
    {\small The tail and head cluster descriptors ($TD_{C_i}$ and $HD_{C_j}$) are extracted from the points' within a time window $\Delta t$ as shown in (a), (b), and (c). The green colored intersection over union (IOU) is calculated as shown in (d) to check if clusters $C_i$ and $C_j$ are a match. The best IOU score between clusters in proximity confirms a match between two clusters, given the score goes over a certain threshold IOU value.} 
    \label{fig:head and tail descriptor}
\end{figure}

We incorporate a mutual information-based descriptor matching method as shown in Fig. \ref{fig:head and tail descriptor}. Suppose we have two clusters, $C_i$ and $C_j$, from the same edge, with cluster $C_i$ preceding cluster $C_j$. Then, the tail descriptor of cluster $C_i$ should match the head descriptor of cluster $C_j$ to be a match. For a short time interval ($\Delta t$), the tail pixel locations of cluster $C_i$ and head pixel locations of cluster $C_j$ are collected as sets, $TD_{C_i}$ and $HD_{C_j}$, where $TD$ and $HD$ stand for tail descriptor and head descriptor, respectively. With $C_i = \{e_k \}^{N}_{k=0}$, the head descriptor ($HD$) is defined as
\begin{equation}
    {HD}_{C_j} = \{\{x_k,y_k\}| t_0 < t_k < t_0 + \Delta t \}
\end{equation}
and the tail descriptor ($TD$) is defined as 
\begin{equation}
    {TD}_{C_i} = \{\{x_k,y_k\}| t_N - \Delta t < t_k < t_N \} .
\end{equation}
The mutual information, $I(C_1,C_2)$, also called intersection over union (IoU), is defined as
\begin{equation}
    I(C_i,C_j) = \frac{TD_{C_i} \cap HD_{C_j}}{TD_{C_i} \cup HD_{C_j}}
\end{equation}
Given the fact that events from the same feature are created in proximity, the descriptor matching is only performed for close instances of the tail of a preceding cluster and the head of a succeeding cluster. The clusters of the largest mutual information are recognized as the same feature and grouped together for longer tracking. As shown in Fig. \ref{fig:3DVisualization}(c), the corresponding clusters are grouped together, enabling longer tracks.

\subsection{Feature Track Extraction}\label{sec:extract}

The feature tracks are estimated by calculating the center of gravity of events in a temporal window. Let cluster $C = \{e_k\}_{k=0}^N$, where $e_k$ is in chronologically ascending order. The feature tracks are extracted by calculating the moving average (MA) within a time window $t_w$. The moving average of event $e_k$, $MA_{t_k}$ is calculated by first collecting the set of events within the time interval $t_w$ at time $t_k$ as:
\begin{equation}
    S_{t_k} = \{e_i|t_k -0.5 t_w < t_i < t_k + 0.5 t_w\} \in C.
\end{equation}
Then, 
\begin{equation}
    MA_{t_k} = \frac{\sum^{M}_{i=0} \{(x_i,y_i)| (x_i,y_i) \in e_i \in S_{t_k}\}}{M},
\end{equation}
where $M$ is the number of elements in $S_{t_k}$.


The window could progress with each event, letting the feature tracks be as fast as the events are detected, which is essentially extracting the feature tracks at very high frequencies.

\section{EVALUATION}
\subsection{Dataset and Parameter Settings}

We evaluated our algorithm with the public event-based camera dataset \cite{mueggler2017event_dataset}. This dataset was recorded with a Dynamic and Active-pixel Vision Sensor (DAVIS) \cite{DAVIS}, with a 240$\times$180 resolution, 130 dB dynamic range, and 3 $\mu$s latency. The dataset also provides ground truth pose of the camera and intensity images like standard cameras. For this dataset, we test our algorithm against 3 different motions (\verb|6dof|, \verb|rotation|, \verb|translation|) in 3 different scenes (\verb|shapes|, \verb|poster|, \verb|boxes|).


The parameters for our method were chosen based on trial-and-error. The parameters are shown in Table \ref{table:run_parameters}.

\begin{table}[]
    \centering
    \captionsetup{font=footnotesize}
    \caption[]{Parameter settings for eCDT}
    
        \begin{tabular}{|l|c|}
        \hline
        k-NN points           & 30    \\ \hline
        Minimum purity score ($\Phi_{min}$) & 0.90  \\ \hline
        Minimum feature age  & 0.01  \\ \hline
        Time window ($t_w$)      & 0.01  \\ \hline
        Search time (s)      & 0.2   \\ \hline
        Threshold Intersection over Union (IoU)   & 0.7 \\ \hline
        \end{tabular}

    \label{table:run_parameters}
\end{table}

\subsection{Comparison with State-of-the-Art Methods}
We compare our method with a previously published open-source tracker: multi-Hypothesis Asynchronous Speed-up Tracking of Events (HASTE) \cite{HASTE4}. While studies such as \cite{ACE2,wacv2020} are also directly related to our work, public implementations are not provided. In contrast, \cite{icra2017zhu} is publicly available, but as HASTE outperforms their study, we compare our results only to HASTE. We use the publicly available implementations provided by the authors which will serve as a baseline to compare our algorithm. 

\subsection{Evaluation Method}
There are two main methods of evaluating the feature tracks: KLT and 3D reprojection. The KLT method was suggested by \cite{gehrig2020eklt}. Since the Event Camera Dataset \cite{mueggler2017event_dataset} provides greyscale images, some works \cite{gehrig2020eklt, wacv2020} utilize these images to compute the ground truth tracks of the feature detection and tracking algorithms. Initialized features are tracked in the images by the Lucas-Kanade tracking algorithm \cite{KLT} with a window size of 35 which is sufficiently large enough to track features of even non-parametric shapes. Since we are using images as a metric for ground truth evaluation, motion blur due to fast movements becomes a threat to this evaluation method. 
Therefore, this method is not utilized for our evaluation. The alternative evaluation method adopted by \cite{HASTE4,ACE2,SILC, Evaluation_2021_journal} evaluates feature tracks by triangulating the 3D point of a feature track with ground truth camera poses obtained by the OptiTrack system at 200Hz. Once the 3D point is estimated, it is reprojected to each ground truth camera pose to calculate the reprojeciton error between the estimated feature point and the tracked feature point. Since this evaluation metric does not rely on the greyscale image, it can even evaluate tracks with motion blur and therefore assess all tracks from the dataset. We report the root mean squared error (RMSE) of each feature track and its feature age - the time elapsed between a feature's initialization and its disposal.  

Given the ground truth camera poses at 200 Hz frequency, we evaluate the 3D reprojection error, which is calculated by triangulating the 3D position of a feature with its track. Let the tracks of feature $i$ be, 
\begin{equation}
    track_i = \{p_0, p_1, ... , p_N\},
\end{equation}
where $p_i$ = ($x_i$, $y_i$, $t_i$) is the spatio-temporal position, and $N$ is the number of feature points in the feature tracks. With the corresponding ground truth poses to each timestamp in $track_i$ as
\begin{equation}
    \boldsymbol{\chi} = \{\chi_0, \chi_1,  ... , \chi_N\},
\end{equation}
where $\chi = (x, y, z, qx, qy, qz, qw)$ is the translation and rotation information of the camera at the corresponding timestamp of $track_i$, triangulating the 3D position of the feature is the same as finding how consistent the feature is with respect to the motion of the camera pose. So finding the 3D reprojection root mean squared error can be evaluated as follows:
\begin{equation}
    E_{rmse} = \frac{\sqrt{\sum_{i=1}^{N} d(p_i,\hat{p_i})^2}}{N},
\end{equation}
where $d(a, b)$ denotes the Euclidean distance between points $a$ and $b$. $\hat{p_i}$ is the projected 2D point of the estimated 3D point. Finding the 3D point $\hat{P} = (x,y,z)$ is done by minimizing $E_{rmse}$. Also, $\hat{p}$ is the projection of a 3D point $\hat{P}$ to a 2D camera view of pose $R$ and $T$. 

Each dataset is evaluated with an error threshold of 3, 5, and 7 pixels. Additionally, since the proposed method is a complete visual front-end capable of providing features and their tracks, we provide HASTE with the same features detected by our algorithm for a fair comparison. eCDT method without the \textit{HT Matching} module will be denoted as eCDT (w/o HT).


%
%
%
\section{RESULTS AND DISCUSSION}

\subsection{Feature Age}

\begin{figure}[h]
    \captionsetup{font=footnotesize}
    \centering
    
    \begin{subfigure}{0.5\textwidth}
      \centering
      \includegraphics[width=\textwidth]{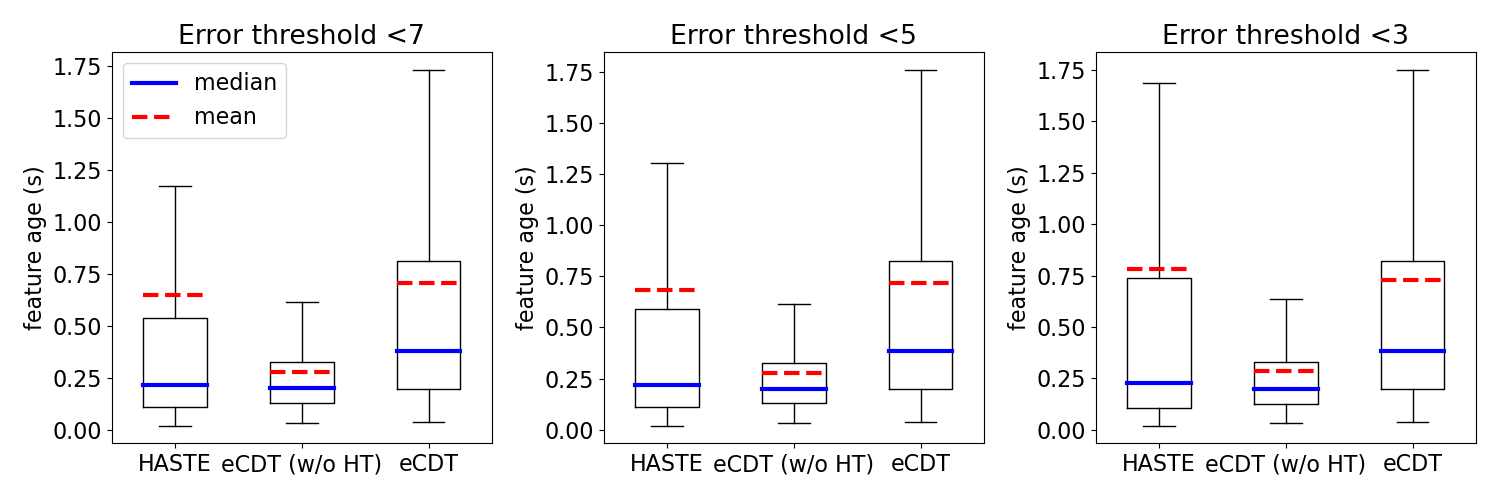}
      \vspace*{-6mm}
      \caption{Feature age of \texttt{boxes}}
    \end{subfigure}
    \vspace*{2mm}
    
    \begin{subfigure}{0.5\textwidth}
      \centering
      \includegraphics[width=\textwidth]{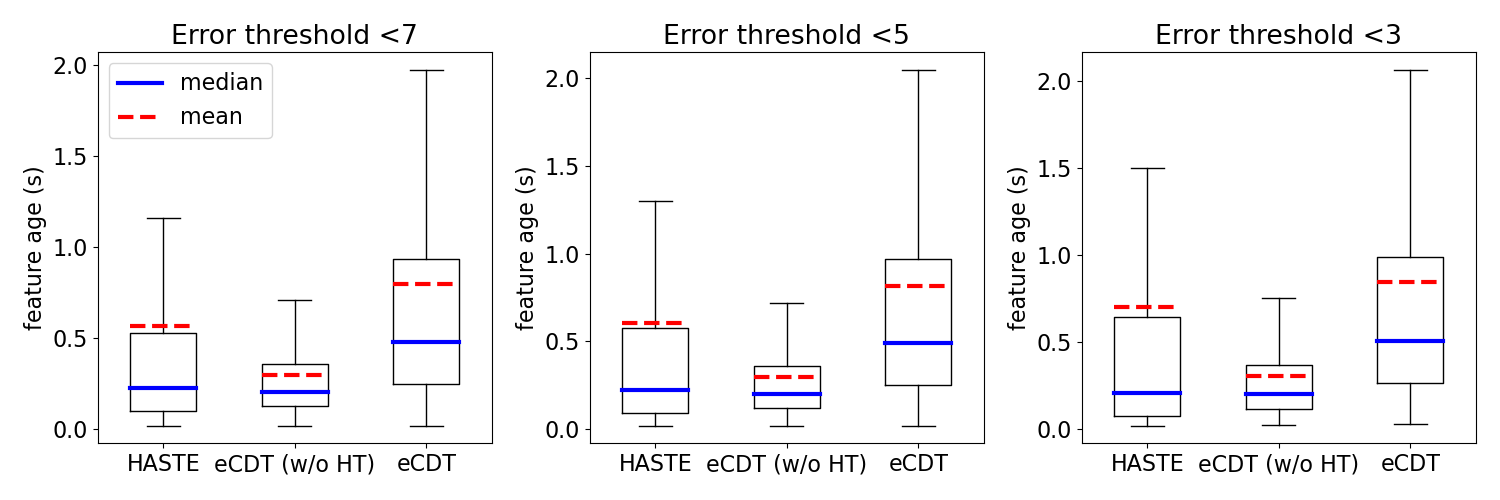}
      \vspace*{-6mm}
      \caption{Feature age of \texttt{poster}}
    \end{subfigure}
    \vspace*{2mm}
    
    \begin{subfigure}{0.5\textwidth}
      \centering
      \includegraphics[width=\textwidth]{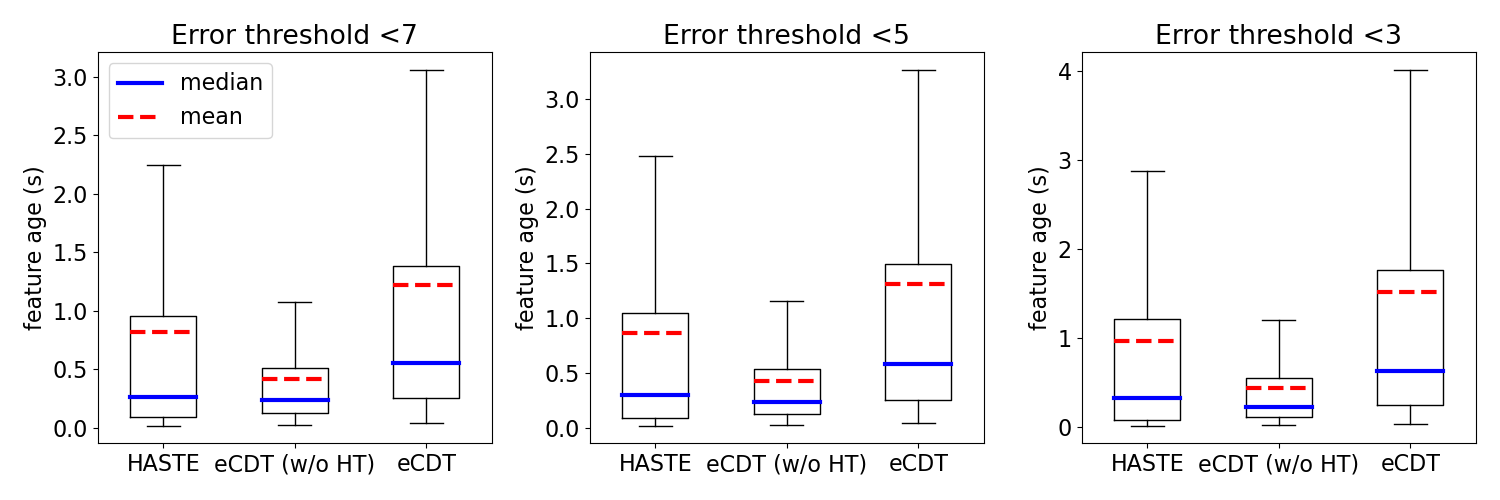}
      \vspace*{-6mm}
      \caption{Feature age of \texttt{shapes}}
    \end{subfigure}
    
    \caption{Boxplots of feature ages for \texttt{boxes}, \texttt{poster}, and \texttt{shapes}. The larger, the better.}
    \label{fig:feature_age_bps}
\end{figure}

\begingroup
\begin{table}[t!]
    \captionsetup{font=footnotesize}
	\centering
	\caption{Comparison with the state of the art method in terms of feature age. Reprojection errors above the threshold values (7, 5, and 3) are considered as outliers and removed. \textit{The more, the better} (unit: s).}  
	{
	\setlength{\tabcolsep}{4pt}
	\begin{tabular}{l|l|cccccc}
	
	\toprule \midrule
	&\multirow{2}[3]{*}{Method} & \multicolumn{2}{c}{7} & \multicolumn{2}{c}{5} & \multicolumn{2}{c}{3} \\  \cmidrule(lr){3-4} \cmidrule(lr){5-6} \cmidrule(lr){7-8} 
	&  & Mean & Med. & Mean & Med. & Mean & Med.   \\ \midrule
	\parbox[t]{2mm}{\multirow{3}{*}{\rotatebox[origin=c]{90}{\texttt{shapes}}}} &HASTE~\cite{HASTE4} & 0.816 & 0.265 & 0.867 & 0.295 & 0.972 & 0.325 \\
	&eCDT w/o HT & 0.417 & 0.240 & 0.427 & 0.235 & 0.446 & 0.227\\
	&eCDT (Ours) & \textbf{1.224} & \textbf{0.550} & \textbf{1.309} & 
	\textbf{0.585} & \textbf{1.518} & \textbf{0.628}  \\ \midrule
	\parbox[t]{2mm}{\multirow{3}{*}{\rotatebox[origin=c]{90}{\texttt{poster}}}} &HASTE~\cite{HASTE4} & 0.564 & 0.225 & 0.606 & 0.220 & 0.699 & 0.205 \\
	&eCDT w/o HT & 0.297 & 0.205 & 0.298 & 0.200 & 0.304 & 0.200\\
	&eCDT (Ours) & \textbf{0.795} & \textbf{0.480} & \textbf{0.814} & \textbf{0.488} & \textbf{0.841} & \textbf{0.505}  \\ \midrule
	\parbox[t]{2mm}{\multirow{3}{*}{\rotatebox[origin=c]{90}{\texttt{boxes}}}} &HASTE~\cite{HASTE4} & 0.648 & 0.215 & 0.684 & 0.220 & \textbf{0.784} & 0.225 \\
	&eCDT w/o HT & 0.275 & 0.200 & 0.277 & 0.200 & 0.283 & 0.200\\
	&eCDT (Ours) & \textbf{0.707} & \textbf{0.380} & \textbf{0.719} & \textbf{0.385} & 0.728 & \textbf{0.385}  \\ \midrule
	\bottomrule
	\end{tabular}
	}
	\label{table:feature_comparison}
\end{table}
\endgroup

\begin{table}[!h]
\captionsetup{font=footnotesize}
\centering
\caption{The average median, mean, and standard deviation of the feature age (s) of each method on 3 scenes, 3 motions, and 3 thresholds, which is a total of 27 samples.}
\begin{tabular}{|c|c|c|c|}
\hline
        & HASTE & eCDT (w/o HT)   & eCDT         \\ \hline
mean (s)    & 0.56  & 0.33          & \textbf{0.92} \\ \hline
median (s) & 0.16  & 0.21          & \textbf{0.46} \\ \hline
stdev (s)   & 0.15  & 0.10          & 0.26          \\ \hline
\end{tabular}

\label{table:Average Mean and std dev feature age}
\end{table}

\begin{table}[!h]
\captionsetup{font=footnotesize}
\centering
\caption{eCDT feature age increase compared to HASTE}
\begin{tabular}{|c|c|c|c|}
\hline
threshold & $<$7    & $<$5   & $<$3    \\ \hline
mean      & 34\%  & 32\% & 26\%  \\ \hline
median    & 100\% & 98\% & 101\% \\ \hline
\end{tabular}
\label{table:percentage better}
\end{table}

\begin{table}[!h]
\captionsetup{font=footnotesize}
\centering
\caption{The paired \textit{t}-Test results comparing the feature age between different methods. \textit{p}-values are presented between two methods on 3 scenes, 3 motions, and 3 thresholds, which is a total of 27 samples.}
\begin{tabular}{|c|c|c|c|}
\hline
\textbf{}               & {HASTE \cite{HASTE4}} & eCDT w/o HT & eCDT (ours)  \\ \hline
HASTE \cite{HASTE4}     & -         & $3.4 \times 10^{-8}$    & $3.8 \times 10^{-7}$                    \\ \hline
eCDT w/o HT              & $3.4 \times 10^{-8}$    & -         & $1.6 \times 10^{-12}$                   \\ \hline
eCDT (ours)             & $3.8 \times 10^{-7}$    & $1.6 \times 10^{-12}$   & -                         \\ \hline
\end{tabular}
\label{table:feature age t-test}
\end{table}

As shown in Fig. \ref{fig:feature_age_bps}, the feature age is longer for eCDT. The detailed numerical results of the feature age are also presented in Table \ref{table:feature_comparison}. For almost all cases, eCDT outperformed HASTE in feature age. Since the feature age does not form a Gaussian distribution, we present both the mean and median feature age of each method in Table \ref{table:Average Mean and std dev feature age}. On average, as presented in Table \ref{table:percentage better}, eCDT's mean and median are 30\% and 100\% longer, respectively. Additionally, by performing the paired \textit{t}-Test between different methods (HASTE, eCDT without \textit{HT matching}, and eCDT), we determine if the feature age difference is significant. The \textit{p}-values between each method are presented in Table \ref{table:feature age t-test}, which shows that eCDT's feature age is significantly better than HASTE's feature age. 

Comparing the results of eCDT with and without \textit{HT matching}, one can notice that the feature age has drastically increased. The feature age has been increased by approximately 200\%, proving that the \textit{HT matching} module is essential for eCDT. The paired \textit{t}-Test between the feature age of eCDT with and without the \textit{HT matching} is $3.4 \times 10^{-8}$, which further highlights the necessity of the process.

\subsection{Reprojection Error}

\begin{table}[!h]
\centering
\captionsetup{font=footnotesize}
\caption{The average median, mean, and standard deviation of the reprojection error (px) of each method on 3 scenes, 3 motions, and 3 thresholds, which is a total of 27 samples. }
\begin{tabular}{|c|c|c|c|}
\hline
        & \multicolumn{1}{c|} {HASTE \cite{HASTE4}} & eCDT w/o HT & eCDT (ours) \\ \hline
median (px)    & \textbf{1.62} & 2.07 & 2.02  \\ \hline
mean (px)    & \textbf{1.83} & 2.21 & 2.17  \\ \hline
stdev (px) & 0.55            & 0.57          & 0.52  \\ \hline
\end{tabular}
\label{table:Average Mean and std dev reprojection error}
\end{table}

\begin{figure}[!h]
    \captionsetup{font=footnotesize}
    \centering
    \begin{subfigure}{0.5\textwidth}
      \centering
      \includegraphics[width=\textwidth]{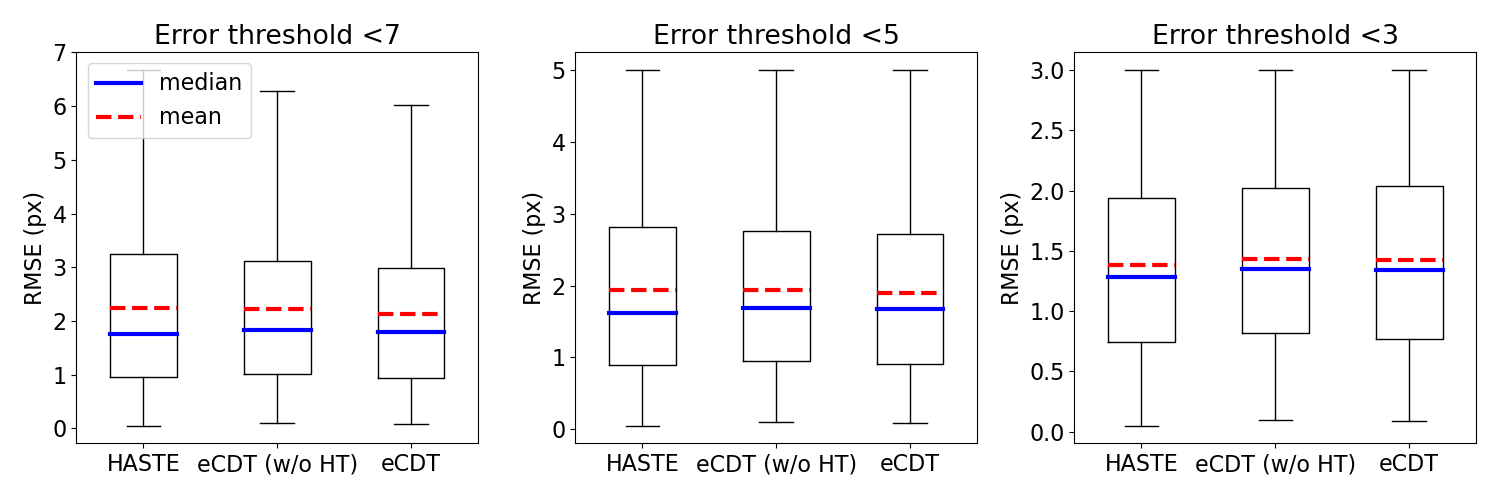}
      \vspace*{-6mm}
      \caption{Reprojection error of \texttt{boxes}}
    \end{subfigure}
    \vspace*{2mm}
    
    \begin{subfigure}{0.5\textwidth}
      \centering
      \includegraphics[width=\textwidth]{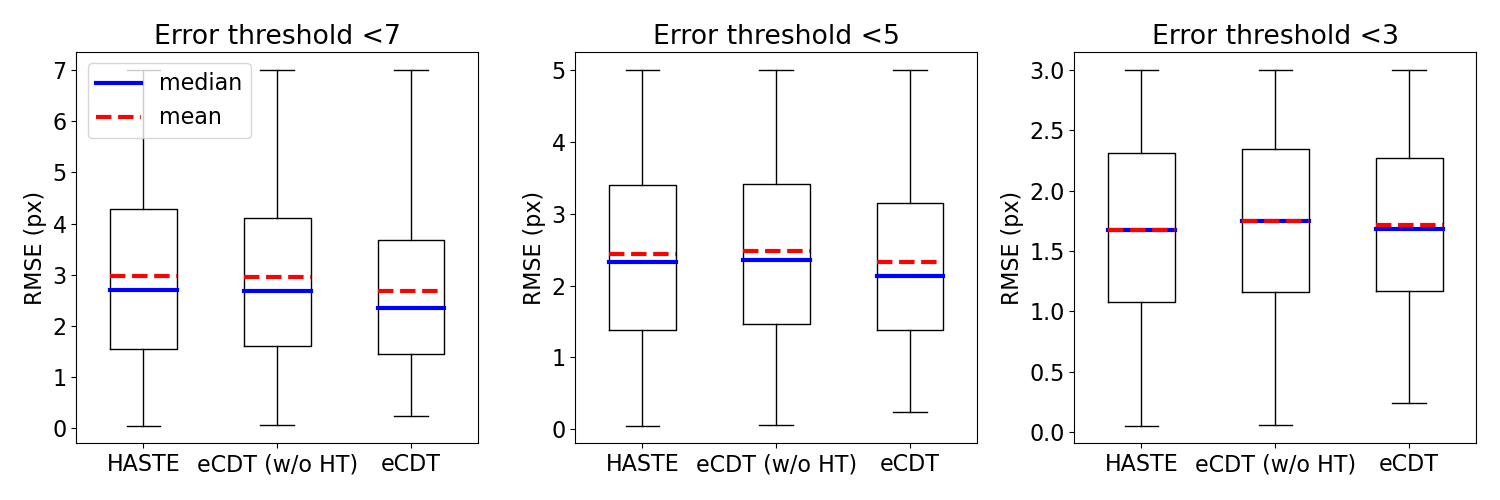}
      \vspace*{-6mm}
      \caption{Reprojection error of \texttt{poster}}
    \end{subfigure}
    \vspace*{2mm}
    
    \begin{subfigure}{0.5\textwidth}
      \centering
      \includegraphics[width=\textwidth]{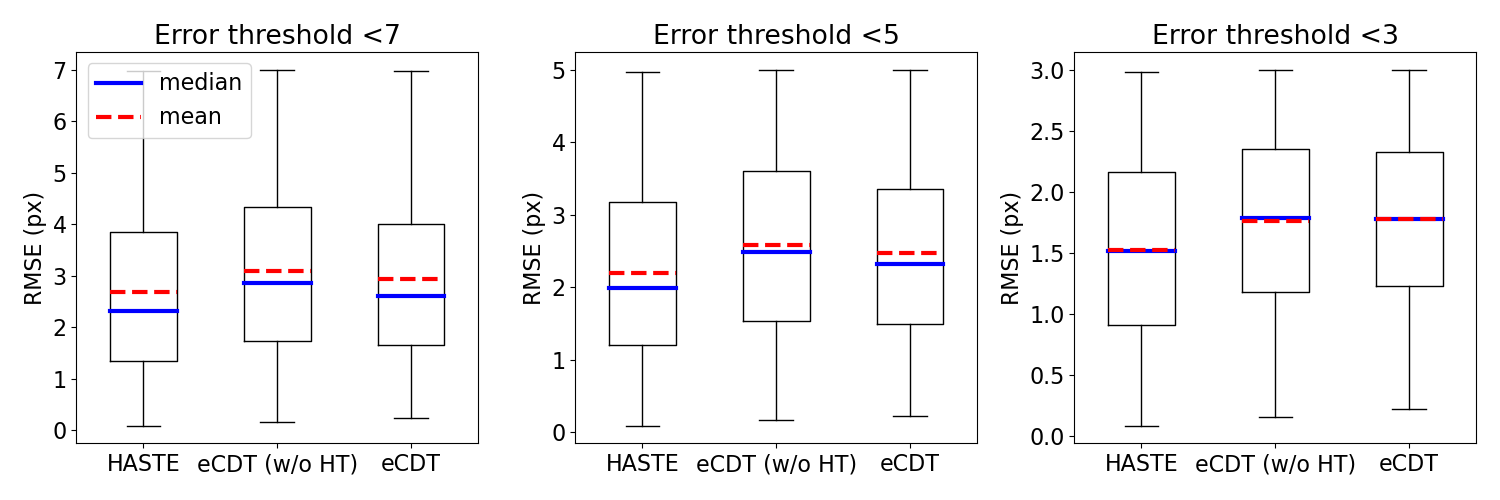}
      \vspace*{-6mm}
      \caption{Reprojection error of \texttt{shapes}}
    \end{subfigure}
    
    \caption{Boxplots of reprojection errors for \texttt{boxes}, \texttt{poster}, and \texttt{shapes}. The smaller, the better.}
    \label{fig:reprojection_error_bps}
\end{figure}

\begingroup
\begin{table}[t!]
    \captionsetup{font=footnotesize}
	\centering
	\caption{Comparison with the state-of-the-art method in terms of reprojection error. Reprojection errors above the threshold values (7, 5, and 3) are considered as outliers and removed. \textit{The less, the better} (unit: RMSE).}  
	{
	\begin{tabular}{l|l|ccc}
	\toprule \midrule
	&Method & 7 & 5 & 3 \\  \midrule
	\parbox[t]{2mm}{\multirow{3}{*}{\rotatebox[origin=c]{90}{\texttt{shapes}}}} &HASTE~\cite{HASTE4} & \textbf{2.68} $\pm$ 1.74 & \textbf{2.19} $\pm$ 1.28 & \textbf{1.53} $\pm$ 0.77 \\
	&eCDT w/o HT & 3.10 $\pm$ 1.67 & 2.58 $\pm$ 1.25 & 1.76 $\pm$ 0.76 \\
	&eCDT (Ours) & 2.94 $\pm$ 1.61 & 2.47 $\pm$ 2.32 & 1.78 $\pm$ 0.67  \\ \midrule 
	\parbox[t]{2mm}{\multirow{3}{*}{\rotatebox[origin=c]{90}{\texttt{poster}}}} & HASTE~\cite{HASTE4} & 2.98 $\pm$ 1.73 & 2.43 $\pm$ 1.27 & \textbf{1.67} $\pm$ 0.76 \\
	&eCDT w/o HT & 2.95 $\pm$ 1.65 & 2.47 $\pm$ 1.22 & 1.75 $\pm$ 0.72 \\
	&eCDT (Ours) & \textbf{2.69} $\pm$ 1.55 & \textbf{2.32} $\pm$ 1.17 & 1.71 $\pm$ 0.69  \\ \midrule 
	\parbox[t]{2mm}{\multirow{3}{*}{\rotatebox[origin=c]{90}{\texttt{boxes}}}} &HASTE~\cite{HASTE4} & 2.24 $\pm$ 1.62 & 1.93 $\pm$ 1.27 & \textbf{1.38} $\pm$ 0.76 \\
	&eCDT w/o HT & 2.22 $\pm$ 1.54 & 1.94 $\pm$ 1.21 & 1.44 $\pm$ 0.74 \\
	&eCDT (Ours) & \textbf{2.12} $\pm$ 1.48 & \textbf{1.90} $\pm$ 1.20 & 1.42 $\pm$ 0.77  \\ \midrule \bottomrule
	\end{tabular}
	}
	\label{table:reproj_comparison}
\end{table}
\endgroup

\begin{table}[!h]
\captionsetup{font=footnotesize}
\centering
\caption{The paired \textit{t}-Test results comparing the reprojection error between different methods. \textit{p}-values are presented between two methods on 3 scenes, 3 motions, and 3 thresholds, which is a total of 27 samples.}
\begin{tabular}{|c|c|c|c|}
\hline
\textbf{} & {HASTE \cite{HASTE4}} & eCDT w/o HT & eCDT (ours)\\ \hline
HASTE \cite{HASTE4}      & -       & 0.01   & 0.02   \\ \hline
eCDT w/o HT              & 0.01   & -       & 0.77   \\ \hline
eCDT (ours)              & 0.02   & 0.77    & -     \\ \hline
\end{tabular}

\label{table:reprojection t-test}
\end{table}

By averaging all scenes, motions, and thresholds, the mean reprojection error is lowest for HASTE with 1.83 px whereas that of eCDT is 2.17 px, which is 0.34 px difference as shown in Table \ref{table:Average Mean and std dev reprojection error}. However, in Fig. \ref{fig:reprojection_error_bps}, the results do not exhibit any consensus towards one's superiority over another since the RMSE values are smaller for HASTE in 5 cases and smaller for eCDT for 4 cases as shown in Table \ref{table:reproj_comparison}. We performed a paired \textit{t}-Test to evaluate the mean reprojection errors between different methods. The \textit{p}-value of the paired \textit{t}-Test is presented in Table \ref{table:reprojection t-test}. While eCDT w/o HT and eCDT has a \textit{p}-value of 0.77, signaling no significant difference between the two methods, HASTE has \textit{p}-values of 0.01 and 0.02 compared to eCDT w/o HT and eCDT, respectively. While these \textit{p}-values represent a statistically significant difference in reprojection error, they are not as significant as the \textit{p}-values from the feature age as shown in Table \ref{table:feature age t-test}. While the reprojection error's \textit{p}-value is in the order of -2, the feature age's \textit{p}-value is in the order of -7, which is different by 5 orders.

Further analyzing the results, considering the fact that HASTE tracks a center point of an $n \times n$ patch while eCDT tracks an edge feature, HASTE should be superior in that it is tracking a more strict feature. When the error threshold is set to 3 pixels, HASTE shows lower reprojection errors. As HASTE tracks point features, it can inherently achieve a lower reprojection error in the lower bound of the spectrum. HASTE also shows better performance when the scene is \verb|shapes|. eCDT for the \verb|shapes| scene generates large edge features which are naturally prone to larger reprojection error since the final feature track is an average point of the large edge feature. Despite these disadvantageous aspects, interestingly, eCDT shows better performance for 4 out of 9 cases, which shows that eCDT tracks features accurately despite tracking edge features. 

\subsection{Quality of Feature Tracking}


\begin{figure}[ht]
\captionsetup{font=footnotesize}
\centering
\begin{subfigure}{0.155\textwidth}
  \centering
  \includegraphics[width=\textwidth]{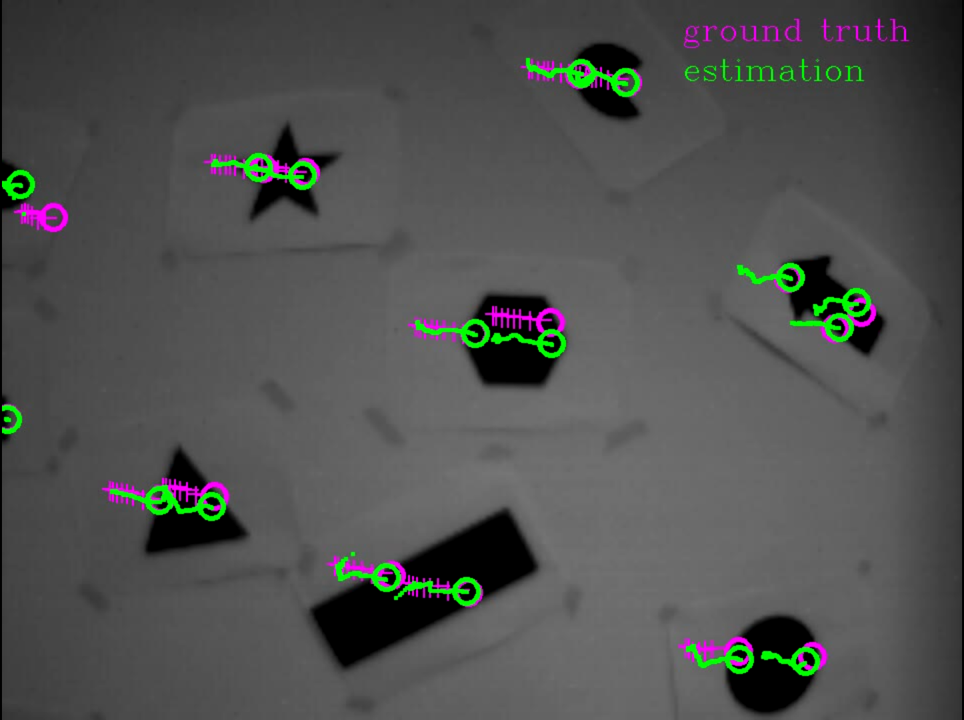}
  \caption{\texttt{shapes}}
\end{subfigure}
\begin{subfigure}{0.155\textwidth}
  \centering
  \includegraphics[width=\textwidth]{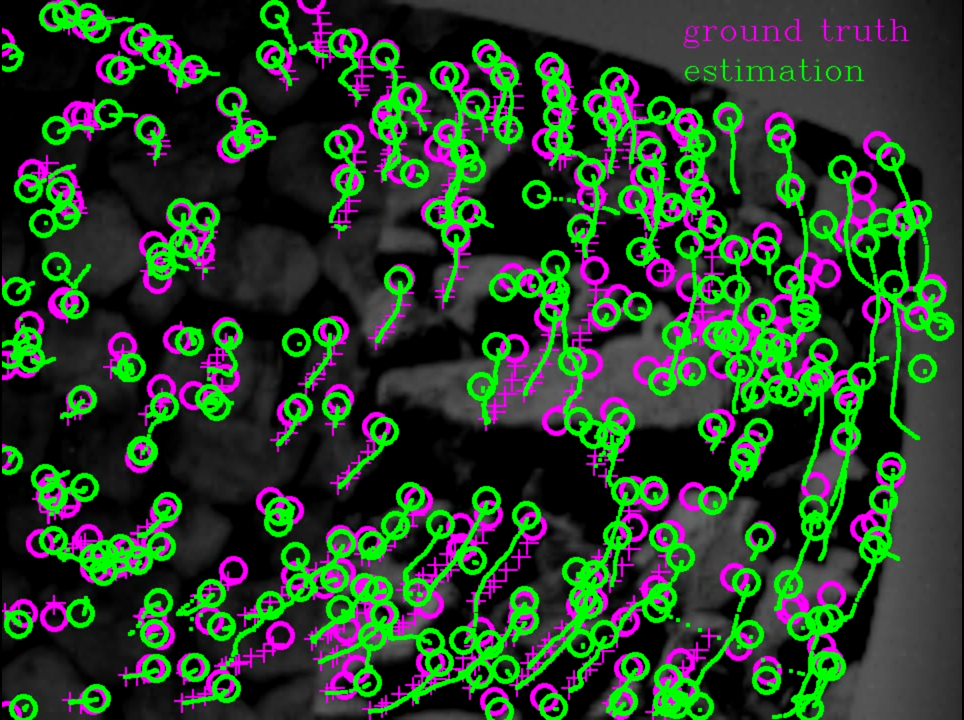}
  \caption{\texttt{poster}}
\end{subfigure}
\begin{subfigure}{0.155\textwidth}
  \centering
  \includegraphics[width=\textwidth]{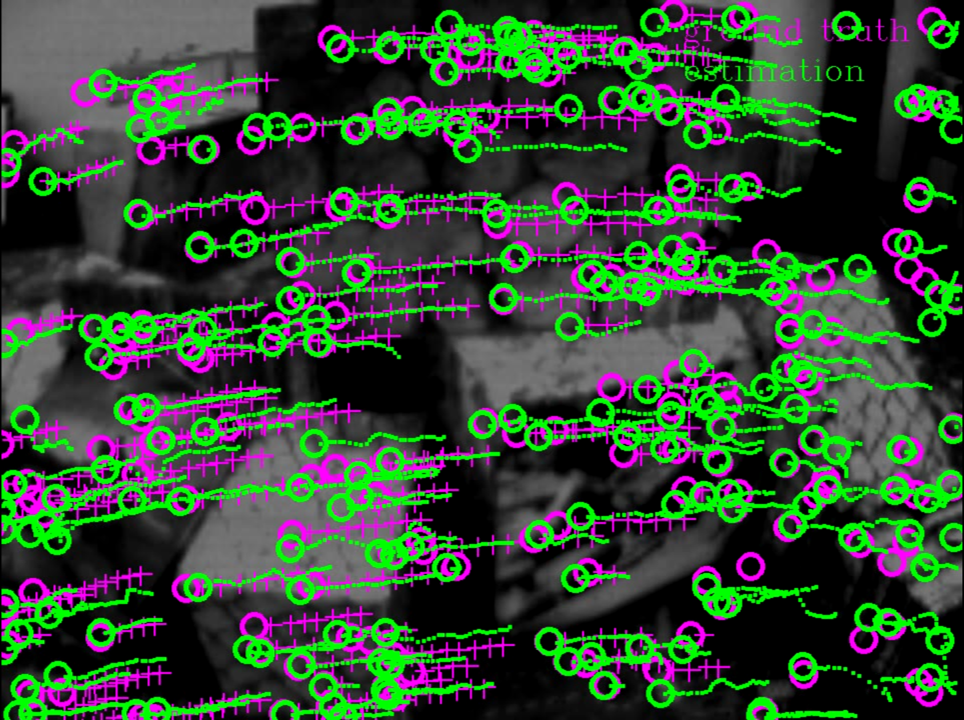}
  \caption{\texttt{boxes}}
\end{subfigure}
\caption[]{Example tracking results of eCDT from \texttt{shapes}, \texttt{poster}, and \texttt{boxes} respectively. The purple-colored ground truth tracks are generated by KLT just for reference.}
\label{fig:vid_example}
\end{figure}

\begin{figure}[ht]
\captionsetup{font=footnotesize}
\centering
\begin{subfigure}{0.235\textwidth}
  \centering
  \includegraphics[width=\textwidth]{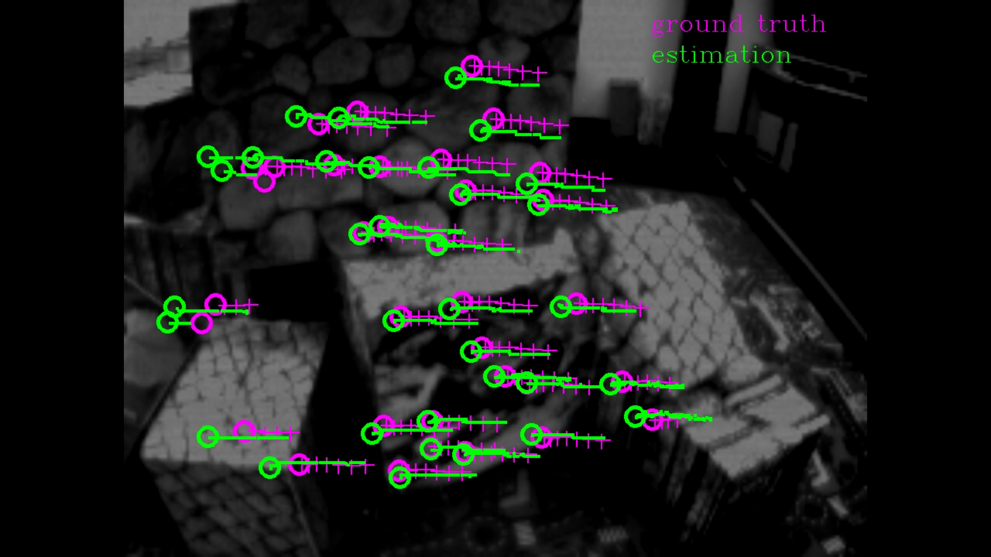}
  \caption{HASTE}
  \label{fig:ex_HASTE} 
\end{subfigure}
\begin{subfigure}{0.235\textwidth}
  \centering
  \includegraphics[width=\textwidth]{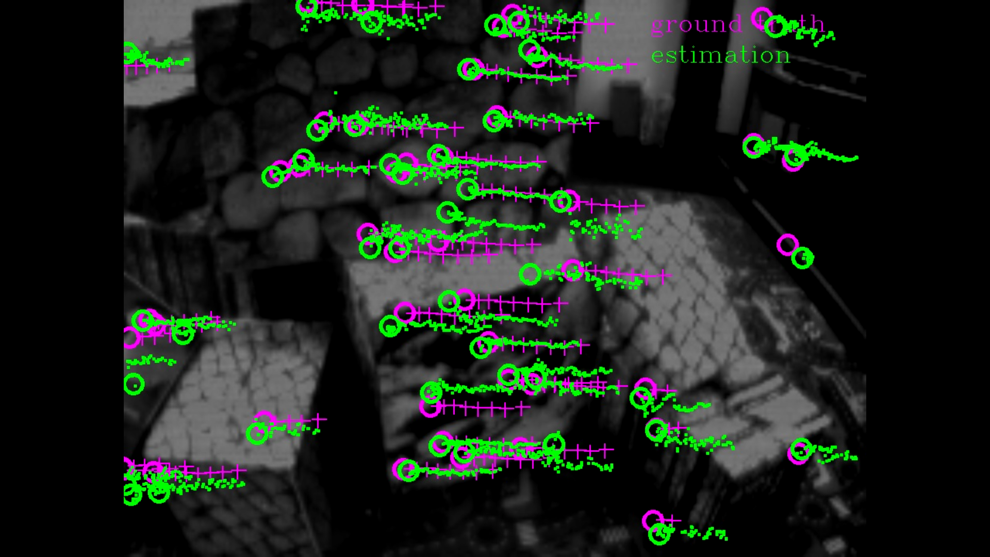}
  \caption{eCDT}
  \label{fig:ex_ecdtstar}
\end{subfigure}
\caption[\texttt{boxes} scene feature tracks for (a) HASTE and (b) eCDT]{\texttt{boxes} scene feature tracks for (a) HASTE and (b) eCDT. The tracking results show that while HASTE is steady, eCDT has noisy tracks. The purple-colored ground truth tracks are  generated by KLT just for reference.}
\label{fig:shaky_tracks}
\end{figure}

Fig. \ref{fig:vid_example} presents the tracking result of eCDT on top of greyscale images to check whether the tracks are visually accurate according to the motion of the camera. Assessing Fig. \ref{fig:shaky_tracks}, we notice that the tracks from eCDT are noisy compared to those from HASTE. Although the features tracked from eCDT are edge features, when the feature tracks are generated, the edge feature tracks are simplified as a single centroid point. Due to this aspect, eCDT's feature point are more susceptible to noise and, therefore, eCDT's tracks are somewhat noisy. 


    




\section{CONCLUSION AND FUTURE WORKS}
In this study, we introduce an entirely novel method of feature detection and tracking for event cameras, which builds upon a key observation that spatio-temporal events are separated by opposing polarity events. By implementing our idea via a novel clustering method, we obtained a simultaneous detection and tracking algorithm. eCDT showed 30\% longer feature tracks on average. Additionally, eCDT showed low reprojection errors on par with the state-of-the-art algorithm, despite tracking edge features. Lastly, the proposed method shows that there lay potential in utilizing the raw events as they are, i.e. without leveraging frame-like representations. We hope this approach functions as a stepping stone to new possible algorithms that utilize raw events.

In the future, we plan to implement this algorithm to be run in real-time for further application. Furthermore, since the motion trajectory of each cluster can be individually extracted, this method has potential for motion segmentation, where dynamic objects are present in the view and segmenting them is the task. 






\section*{ACKNOWLEDGMENT}

This research was supported by BK21 FOUR and StradVision. We appreciate all the supports of StradVision members who provided insight and expertise. The contents are solely the responsibility of the authors.




\bibliographystyle{./IEEEtran}
\bibliography{./iros22,./IEEEabrv}

\addtolength{\textheight}{-12cm}

\end{document}